\pdfoutput=1

\documentclass[11pt]{article}

\usepackage[final]{acl}
\usepackage{amsmath} 
\usepackage{breqn}
\usepackage{amsfonts} 
\usepackage{float}
\usepackage{titlesec}
\usepackage{placeins}
\usepackage{caption}
\usepackage{tcolorbox}

\usepackage{times}
\usepackage{latexsym}
\usepackage{multirow}
\usepackage[shortlabels]{enumitem} 
\usepackage{bm}
\usepackage{mathrsfs,amssymb}

\usepackage[T1]{fontenc}

\usepackage[utf8]{inputenc}
\usepackage{makecell}
\usepackage{microtype}

\usepackage{inconsolata}
\usepackage{enumitem}

\usepackage{graphicx}
\usepackage{subcaption}

\usepackage{amsmath}
\DeclareMathOperator*{\argmax}{arg\,max}

\graphicspath{{./}{../}}

\title{\textsc{ReCall} \includegraphics[width=0.9cm]{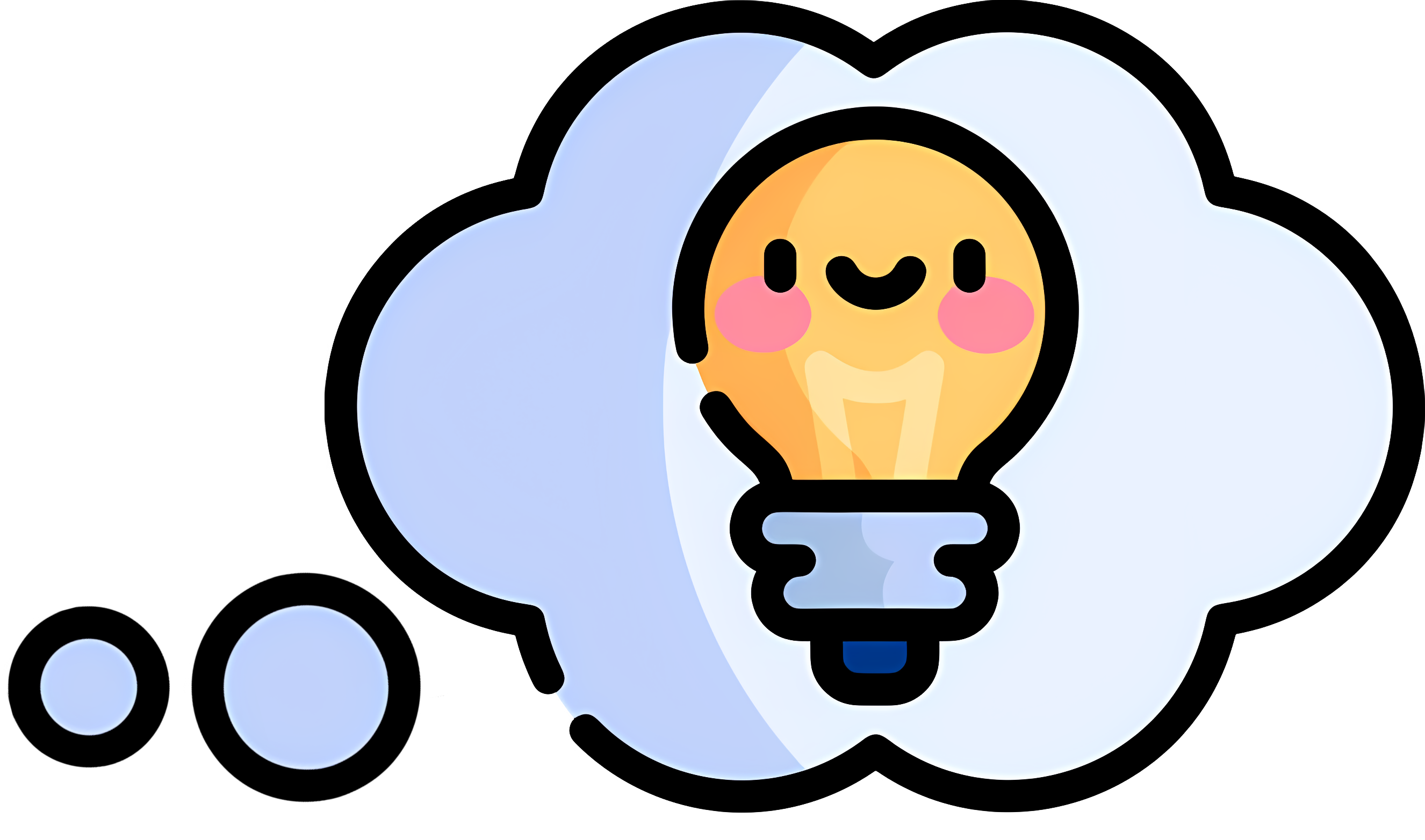}: \\ Library-Like Behavior In Language Models is Enhanced  \\by Self-Referencing Causal Cycles\\}

\author{
\textbf{Munachiso Nwadike$^{1,2}$, Zangir Iklassov$^{1}$, Toluwani Aremu$^{1}$, } \\
\textbf{Tatsuya Hiraoka$^{1,2}$, Benjamin Heinzerling$^{2,3}$, Velibor Bojkovic$^{1}$,} \\
\textbf{Hilal Alqaubeh$^{1,2}$, Martin Tak\'a\v{c}$^{1}$, Kentaro Inui$^{1,2,3}$} \\ 
$^{1}$MBZUAI, Abu Dhabi, UAE \, • \, $^{2}$RIKEN AIP, Japan \, • \, $^{3}$Tohoku University, Japan \\ 
\texttt{\footnotesize munachiso.nwadike@mbzuai.ac.ae} 
}

\usepackage[textsize=tiny]{todonotes}

\usepackage{booktabs}

\usepackage{amsmath} 

\begin{document}
\maketitle
 
\begin{abstract}
We introduce the concept of the \textit{self-referencing causal cycle} (abbreviated \textsc{ReCall})—a mechanism that enables large language models (LLMs) to bypass the limitations of unidirectional causality, which underlies a phenomenon known as the \textit{reversal curse}. When an LLM is prompted with sequential data, it often fails to recall preceding context. For example, when we ask an LLM to recall the line preceding ``O say does that star-spangled banner yet wave'' in the U.S. National Anthem, it often fails to correctly return ``Gave proof through the night that our flag was still there''—this is due to the reversal curse. It occurs because language models such as ChatGPT and Llama generate text based on preceding tokens, requiring facts to be learned and reproduced in a consistent token order. While the reversal curse is often viewed as a limitation, we offer evidence of an alternative view: it is not always an obstacle in practice. We find that \textsc{ReCall} is driven by what we designate as \textit{cycle tokens}—sequences that connect different parts of the training data, enabling recall of preceding tokens from succeeding ones. Through rigorous probabilistic formalization and controlled experiments, we demonstrate how the cycles they induce influence a model's ability to reproduce information. To facilitate reproducibility, we provide our code and experimental details at \url{https://github.com/samunaai/remember}. 
\end{abstract}


\begin{figure}[!t]
\centering
\includegraphics[width=1.0\linewidth]{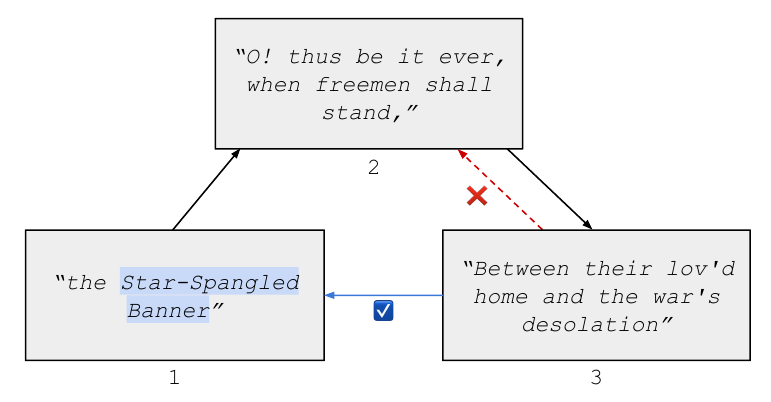}
\caption{In a text containing the U.S. National Anthem, the phrase ``Star-Spangled Banner'' appears several times and functions as a cycle token-sequence, creating a loop termed a \textit{self-referencing causal cycle}. This cycle reconnects later parts of the text to earlier ones. Without these cycle tokens, the model would be unable to predict the preceding line ``O! thus be it ever, when freemen shall stand'' from the succeeding line ``Between their lov'd home and the war's desolation'' due to left-to-right causality (the red line). The blue line indicates that the cycle token-sequence occurs at multiple points in the text, acting as a reference point for the model to retrieve preceding context.}
\label{cycle_diagram_text}
\end{figure}


\section{Introduction}

Consider, by way of metaphor, a large language model  (LLM) as the parametric equivalent of a physical library of knowledge \cite{lederman2024language}. A library evokes structured collections of books or documents, each cataloged for efficient retrieval. Similarly, pretraining LLMs on billions of tokens transforms them into repositories of encoded knowledge \cite{petroni2019language,heinzerling2020language,wang2024knowledge}. Prompts, therefore, act as cross-references, directing retrieval of specific information, much like library indexes facilitate access to books on shelves.

In a library, we expect to retrieve information reliably. However, language models are not always suited for this. For example, when asked to recall the line preceding ``Between their lov'd home and the war's desolation'' in the U.S. National Anthem, a language model often fails to return ``O! thus be it ever, when freemen shall stand'' (see Figure~\ref{cycle_diagram_text}). This is a symptom of the ``reversal curse,'' where models struggle to generalize reversed relationships from their training data.

In this work, we explore a mechanism within LLMs that naturally mitigates the reversal curse leveraging inherently occurring patterns in pretraining data. Specifically, we show how self-referencing causal cycles (\textsc{ReCall}) emerge from repeated token sequences, allowing models to bypass this limitation without requiring explicit in-context reversal strategies. These cycles, induced by what we term \textit{cycle tokens}, act as natural hyperlinks between different parts of a text, enhancing memory retrieval. Our analysis focuses on autoregressive models, for which we propose a novel two-step \textsc{ReCall} process to improve information retrieval.

\section{Related Work}

The reversal curse, or ``inverse search problem,'' has been extensively studied in large language models. Early works by \citet{berglund2023reversal} and \citet{allen2023physics} identify the issue, noting that models struggle to generalize relationships such as ``A after B'' to ``B before A.'' This limitation, stemming from the autoregressive nature of models like GPT \cite{achiam2023gpt} and LLaMA \cite{dubey2024llama}, persists even in those with chain-of-thought capabilities \citep{guo2024mitigating}.

Recent studies highlight how reverse thinking enhances reasoning abilities \citep{chen2024reverse}, leveraging in-context learning to address tasks such as arithmetic and logical deduction. However, it is already known that the reversal curse does not manifest in in-context learning settings \citep{berglund2023reversal}, as models benefit from explicit contextual information during inference.

Bidirectional models, like BERT \citep{devlin2019bert}, avoid the reversal curse by using masked token objectives \citep{wu2024exploring}, allowing them to reason about context in both directions. However, these models are not designed for autoregressive tasks such as next-token prediction, which underpins state-of-the-art chatbots.

Methods to mitigate the reversal curse often involve data augmentation. For example, \citet{guo2024mitigating} and \citet{golovneva2024reverse} explore token permutation techniques, while \citet{springer2024repetition} propose token repetition to enhance causal links in training data. Unlike these manual interventions, our work investigates naturally occurring patterns in pretraining data, and how they organically mitigate the reversal curse.

\section{Formalizing \textsc{ReCall}}
\label{mathematical_rigor_section}
This section formalizes \textsc{ReCall} by introducing the concept of ``cycle tokens'' and their causal effects. To establish this foundation, we begin by revisiting the reversal curse in probabilistic terms.

\subsection{Revisiting the Reversal Curse}
Consider the set of all possible token sequences, denoted by \(\mathcal{S}\), for a given textual dataset that adheres to a true data distribution \(P\). For simplicity, we assume the language is written in a left-to-right (LTR) script, such as English. However, the arguments generalize naturally to other text directions, such as right-to-left (RTL, e.g., Arabic) or top-to-bottom (TTB, e.g., Japanese), requiring only minor notational modifications.

Let \(\mathcal{S}_{\text{seq}} \in \mathcal{S}\) be a sequence of \(n\) tokens, represented as \(\mathcal{S}_{\text{seq}} := [e_1, e_2, \dots, e_n]\). We partition \(\mathcal{S}_{\text{seq}}\) into two segments at some index \(i\), where \(S_l := [e_1, e_2, \dots, e_i]\) represents the left-hand segment, and \(S_r := [e_{i+1}, \dots, e_n]\) represents the right-hand segment. Probabilistically, \(S_r\) is the most likely continuation of \(S_l\) under the distribution \(P\). For example, \(S_r\) could correspond to the continuation \emph{``O say does that star-spangled banner yet wave,''} while \(S_l\) represents the preceding context \emph{``Gave proof through the night that our flag was still there.''} Identifying that the latter precedes the former poses a challenge for autoregressive models. The reason for this can be illustrated as follows:

For an autoregressive model \(\mathcal{M}\) trained on the true data distribution, we have \(P \approx P_{\mathcal{M}}\), where \(P_{\mathcal{M}}\) denotes the model’s learned approximation of the true data distribution. Accordingly, we expect:
\begin{equation}
\label{original_equation}
S_r = \argmax_{s \in \mathcal{S}} P_{\mathcal{M}}(s|S_l).
\end{equation}

This implies that \(\mathcal{M}\) can readily produce the highest-probability right-hand sequence \(S_r\) given a left-hand sequence \(S_l\). However, the model struggles to select the correct \(S_l\) given \(S_r\), because:
\begin{equation}
S_l \neq \argmax_{s \in \mathcal{S}} P_{\mathcal{M}}(s|S_r).
\end{equation}
Instead, the model can only indirectly compute the left-hand sequence using Bayes' rule:
\begin{equation}
\label{explained_in_appendix}
S_l = \argmax_{s \in \mathcal{S}} P_{\mathcal{M}}(S_r|s) P_{\mathcal{M}}(s).
\end{equation}
Equation~(\ref{explained_in_appendix}) is explained in greater detail in Appendix~\ref{background_probability}.

However, computing the argmax in equation \ref{explained_in_appendix} would require iterating through all possible sequences $s \in \mathcal{S}$, pairing each $s$ with the fixed $S_r$, and then evaluating the combined sequence $[s, S_r]$ using the model $\mathcal{M}$ to obtain $P_{\mathcal{M}}(S_r|s)$ and $P_{\mathcal{M}}(s)$.

While the model can easily compute $P_{\mathcal{M}}(S_r|s)$, which represents the likelihood of the right sequence given the left part, and $P_{\mathcal{M}}(s)$, the prior probability of a token sequence, iterating over all possible sequences $s \in \mathcal{S}$ is computationally infeasible due to the combinatorial growth of possible sequences. 

To address this, we would like to narrow the search to a smaller candidate set $S_{l_c}$, generated by prompting the right-hand sequence:
\begin{equation}
S_l = \argmax_{s \in S_{l_c}} P_{\mathcal{M}}(S_r|s) P_{\mathcal{M}}(s).
\end{equation}
However, how do we construct $S_{l_c}$ if $S_r$ offers no direct clues about its preceding sequence? We would require knowledge of the sequence to the left $S_{l}$ (left-of-left), which we do not have.

\subsection{Introducing Cycle Tokens}
\label{formalizing_cycle_tokens}
Our core hypothesis is that, instead of predicting $S_l$ directly from $S_r$, we construct a modified sequence $S_r' := [e_{i+1}, \dots, e_n, e_1]$ by appending $e_1$ to the end of $S_r$. This modified sequence serves as a pointer back to the start of the original sequence, providing access to $S_l$. The token $e_1$ acts as a \textit{cycle token}—so named because it induces a cycle in the causal flow of next-token prediction, enabling the model to effectively “see” left-hand tokens from the right-hand side.

\begin{figure}[!t]
\centering
\includegraphics[width=0.5\linewidth]{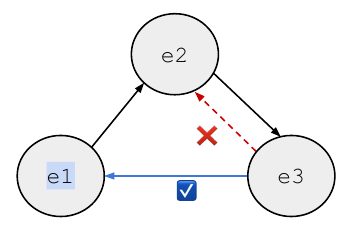}
\caption{A token sequence $[e1, e2, e3]$ illustrates the difficulty of predicting $e2$ from $e3$ due to left-to-right causality. By appending token $e1$ to the end of the sequence, a loop is formed, allowing the model to transition from $e3$ back to $e2$. Red and blue indicate the same concepts as in Figure \ref{cycle_diagram_text}.}

\label{cycle_diagram_token}
\end{figure}

From this modified sequence, we can extract $S_l' := [e_2, e_3, \dots, e_i]$ using continued next-token predictions. Importantly, the entire sequence $S_r$ does not need to be repeated; even a single cycle token can serve as a pointer, creating what we term a \textit{self-referencing causal cycle} (see Figure \ref{cycle_diagram_token}).

As the length $i$ of $S_l$ grows, $S_l'$ increasingly resembles $S_l$. For example, consider two sentences of 100 words each and two sentences of 5 words each. With one differing letter in each pair, the longer sentences exhibit greater similarity.

\noindent Formally:
\[
S_r' := S_r \oplus e_1 \quad \text{and} \quad S_l := e_1 \oplus S_l',
\]
where $\oplus$ denotes concatenation. Therefore:
\begin{equation}
S_l \approx S_l'.
\end{equation}
Noting that:
\begin{equation*}
S_l' \to S_l \quad \text{as} \quad i \to \infty.
\end{equation*}

Thus, in the presence of a self-referencing causal cycle, a left-to-right autoregressive model can approximate left-hand sequence information from the right-hand side. Interpreted intuitively, the self-referencing causal cycle is a mechanism that allows the model to `loop back' and access earlier parts of sequences of any length, thereby introducing a bidirectional influence to a unidirectional model.

\section{Experiments and Analysis} 

\subsection{Deterministic Few-Token \textsc{ReCall}}
\label{fewtoken_section}

To demonstrate self-referencing causal cycles, we use simple few-token datasets and a small decoder-based transformer model \cite{vaswani2017attention} with two layers and eight attention heads. The datasets are designed for ease of interpretation. Implementation details are provided in Appendix \ref{appendix_exp_settings}.

Suppose we train our model on four-token sequences of the form $[e1, e2, e3, e1]$. Let $e1$ be an integer randomly selected from $[1, 100]$, $e2$ from $[101, 200]$, and $e3$ from $[201, 300]$. If each integer is randomly selected, such that each e1 can only be paired with a unique e2, and each e2 with a unique e3, then the dataset will consist of 100 samples. For example, one possible training sample is $\mathcal{S}_{train} = [79, 155, 264, 79]$, where $e1$ appears twice—at the beginning and the end of the sequence. During training, the transformer memorizes this sequence, and we test whether it can predict the sequence $\mathcal{S}_{test} = [264, 79, 155]$. If the model can predict this sequence $\mathcal{S}_{test}$ with 100\% accuracy, it demonstrates that it can recover $155$ from $264$ by using the token $79$ as a ``cycle'' to link the end of the sequence back to its beginning. In this case, token $79$ serves as the cycle token.

In all experiments, we consider transitions from a right-hand \textit{token} (denoted lowercase $e$) to a left-hand \textit{token}, and from one \textit{sequence} (denoted uppercase $E$) to another. For example, instead of memorizing $\mathcal{S}_{\text{train}} = [e1, e2, e3, e1]$, we could memorize $\mathcal{S}_{\text{train}} = [e1, E2, E3, e1]$ and test recovery of $\mathcal{S}_{\text{test}} = [E3, e1, E2]$. Such sequence-to-sequence experiments better reflect real-world pretraining data, where information spans phrases or sentences.

\begin{figure}[H]
\centering
\includegraphics[width=1.0\linewidth]{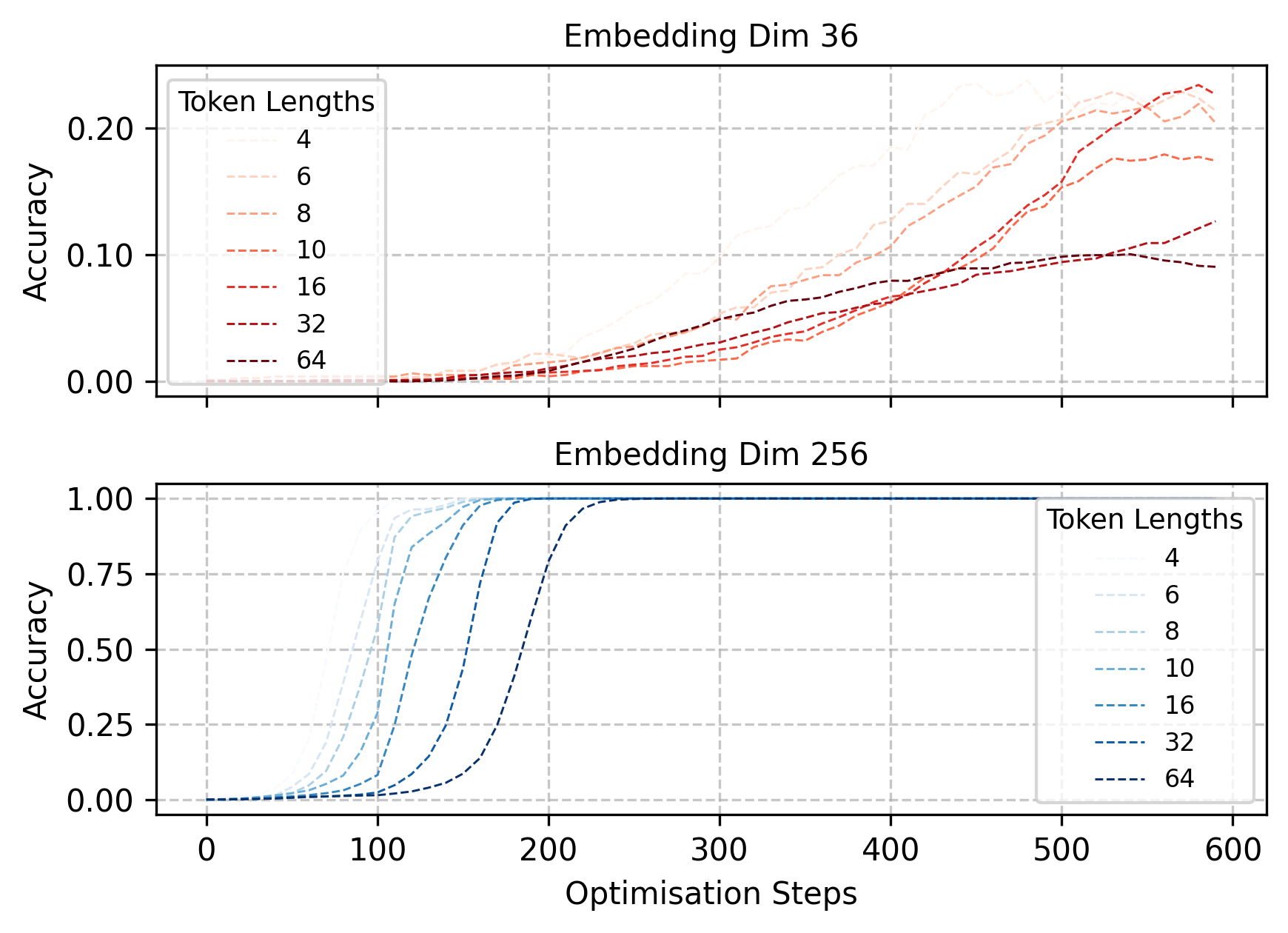}
\caption{Increasing the value of $\mathcal{N}$ results in longer token-sequences, but validation accuracy remains high with sufficient embedding dimensionality.} 
\label{ablation_graph_on_length_of_path}
\end{figure}

\begin{table}[!t]
\centering
\resizebox{0.99\linewidth}{!}{%
\begin{minipage}{0.5\linewidth} 
    \centering
    \small
    \setlength{\tabcolsep}{0.6em} 
    \renewcommand{\arraystretch}{1.6} 

    \resizebox{1.0\linewidth}{!}{
    \begin{tabular}{c c c c} 
    \toprule 
    \multirow{1}{*}{Experiment} & Memorized Sequence & Reversal Path & Viable \\
    \midrule 
    \multirow{2}{*}{Baseline} & (e1, e2, e3, e1) & e3 $\rightarrow$ e1 $\rightarrow$ e2 & \includegraphics[height=1em]{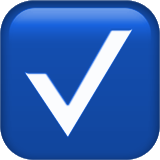} \\
    \cmidrule(lr){2-4}
    & (e1, E2, E3, e1) & E3 $\rightarrow$ e1 $\rightarrow$ E2 & \includegraphics[height=1em]{images/bluecheckmark_thick.png} \\ 
    \midrule 
    \multirow{2}{*}{Length of Path} & (e1, e2, e3, E4, e1) & e3 $\rightarrow$ E4 $\rightarrow$ e1 $\rightarrow$ e2 & \includegraphics[height=1em]{images/bluecheckmark_thick.png} \\ 
    \cmidrule(lr){2-4}
    & (e1, E2, E3, E4, e1) & E3 $\rightarrow$ E4 $\rightarrow$ e1 $\rightarrow$ E2 & \includegraphics[height=1em]{images/bluecheckmark_thick.png} \\ 
    \midrule 
    \multirow{2}{*}{\makecell{Length of \\ `Out-of' Path}} & (e1, e2, E3, e4, e1) & e4 $\rightarrow$ e1 $\rightarrow$ e2 & \includegraphics[height=1em]{images/bluecheckmark_thick.png} \\ 
    \cmidrule(lr){2-4}
    & (e1, E2, E3, E4, e1) & E4 $\rightarrow$ e1 $\rightarrow$ E2 & \includegraphics[height=1em]{images/bluecheckmark_thick.png} \\ 
    \midrule 
    \multirow{2}{*}{\makecell{Cycle \\ Composability}} & \makecell{(e1, e2, e3) and \\ (e3, e1, e4)} & e3 $\rightarrow$ e1 $\rightarrow$ e2 & \includegraphics[height=1em]{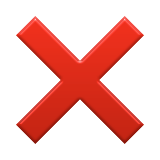} \\  
    \cmidrule(lr){2-4}
    & \makecell{(e1, E2, E3) and \\ (E3, e1, E4)} & E3 $\rightarrow$ e1 $\rightarrow$ E2 & \includegraphics[height=1em]{images/redxmark_thick.png} \\  
    \midrule 
    \multirow{2}{*}{\makecell{Hyperlink \\ Composability}} & \makecell{(e5, e3, e1, e4) and \\ (e0, e1, e2, e3)} & e2 $\rightarrow$ e3 $\rightarrow$ e1 $\rightarrow$ e4 & \includegraphics[height=1em]{images/bluecheckmark_thick.png} \\ 
    \cmidrule(lr){2-4}
    & \makecell{(E5, E3, e1, E4) and \\ (E0, e1, E2, E3)} & E2 $\rightarrow$ E3 $\rightarrow$ e1 $\rightarrow$ E4 & \includegraphics[height=1em]{images/bluecheckmark_thick.png} \\ 
    \bottomrule  
    \end{tabular}}
\end{minipage}
}
\caption{List of the few-token experiments showing the sequences, causal paths, and the viability of reversal. These experiments explore various configurations, including baseline setups, variations in path length, out-of-path noise, and cross-sample hyperlinks, to evaluate the effectiveness of cycle tokens in linking sequences. $e1$ is the cycle token in all cases, with the exception of \texttt{Hyperlink Composability}, where $e3$ and $E3$ are the respective cycle token and token-sequences.}
\label{table_for_few_token}
\end{table}
The settings described above form our \texttt{Baseline} few-token experiment, outlined in Table \ref{table_for_few_token}. We introduce several variations on this experiment to test the robustness of self-referencing causal cycles. For instance, instead of transitioning directly from $e3$ to $e1$, we can introduce a sequence of ``noise'' tokens $E4$ in between. If $E4$ has a length $\mathcal{N}=3$, there are three tokens blocking the transition path from the start token $e3$ to the cycle token $e1$. We term this the 
\texttt{Length of Path} experiment, as we vary the length $\mathcal{N}$ (see Figure \ref{ablation_graph_on_length_of_path}). Alternatively, we could include a spurious sequence $E3$ between two tokens, $e4$ and $e2$, in the right-to-left direction but not in the left-to-right direction (i.e., not in the transition path from $e4$ to the cycle token). We term this the \texttt{Length of `Out-of' Path} experiment. If the model memorizes all such out-of-path sequences with perfect validation accuracy, it reinforces the evidence that $e1$ is responsible for creating the self-referencing causal cycle.

Another characteristic of real-world data is that token-sequences carrying related information may be distributed across samples in separate data batches. The \texttt{Hyperlink Composability} experiment demonstrates that a cycle token can hyperlink different samples, enabling causal ``jumps'' between them. For instance, if we have two samples in different batches, $\mathcal{S}_{1} = [e5, e3, e1, e2]$ and $\mathcal{S}_{2} = [e0, e1, e2, e3]$, we can ``jump backwards'' and recover $\mathcal{S}_{test} = [e2, e3, e1, e2]$ where $e3$ acts as a bridge token.

\begin{figure}[H]
\centering
\includegraphics[width=\linewidth]{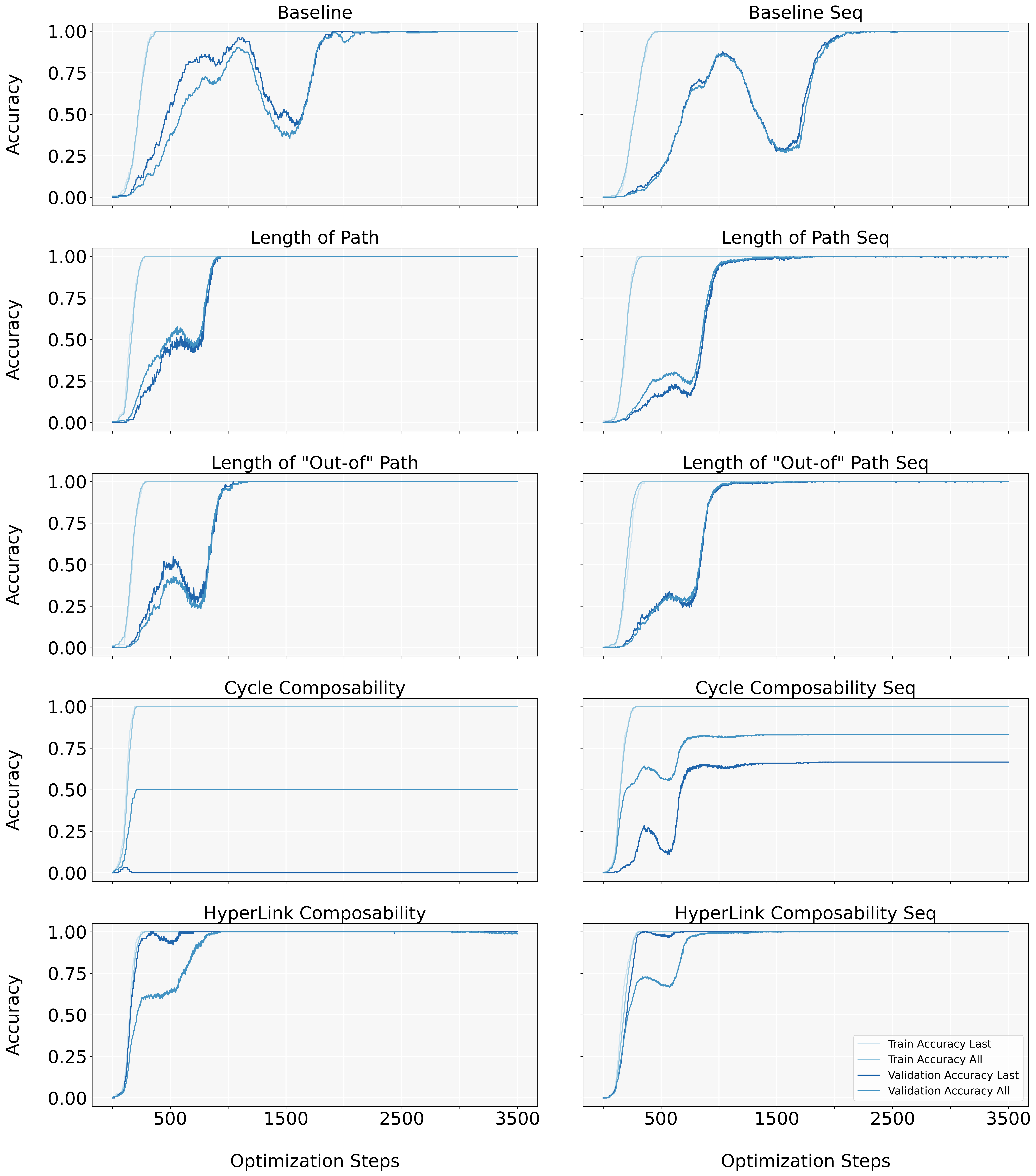}
\caption{Performance of cycle tokens in few-token experiments, showing validation accuracy as a function of optimization steps (epochs). Train-validation splits follow Figure~\ref{table_for_few_token}. Cycle tokens enable transitions from the starting to the target token-sequence, bypassing left-to-right causality. \texttt{Last} accuracy refers to correct prediction of the final token-sequence---the base case of which is a single token---while \texttt{All} accuracy refers to correct prediction of all subsequent tokens after the first. For example, in a validation sequence $(e_3, e_1, e_2)$, \texttt{All} means correctly predicting $e_1$ and $e_2$, while \texttt{Last} refers to predicting $e_2$. In a mixed-sequence case like $(E_3, e_1, E_2)$, \texttt{All} requires correctly predicting both $e_1$ and $E_2$, and \texttt{Last} refers to the correct prediction of $E_2$. In all scenarios, 100\% accuracy is achieved in predicting the left-hand sequence using the cycle token as a bridge.}
\label{few_token_10_experiments}
\end{figure}

\textbf{\textsc{ReCall} is achievable \textbullet} In nearly all cases, we observe perfect generalization after training for a specified number of steps, as shown in Figure \ref{few_token_10_experiments}.  The sole exception is the \texttt{Cycle Composability} experiment, which reveals that cycle tokens function more effectively when the left-hand context does not alter their semantics. For example, consider a training dataset consisting of $\mathcal{S}_{1} = [e1, e2, e3]$ and $\mathcal{S}_{2} = [e3, e1, e4]$, and we seek to transition from  $e3$ to $e2$ (i.e., $\mathcal{S}_{test} = [e3, e1, e2]$).  If $e3$ were not present in  $\mathcal{S}_{2}$, a transition from $e1$ to $e2$ would be possible by virtue of \texttt{Hyperlink Composability}. In this case, the model would have a 50\% chance of predicting $e2$ following $e1$ (a stochastic setting discussed further in Section \ref{stochastic_experiments_section}). However, because ${\mathcal{S}}_{2}$ is present in the training data, the model will be inclined to predict $e4$ after seeing $[e3,e1]$ 100\% of the time, as $[e3, e1, e4]$ represents a pattern trained into it via cross entropy loss. This behavior is desirable, as it demonstrates how self-attention naturally alters the semantic interpretation of a token based on preceding context.

\textbf{\textsc{Recall} is consistent over varying sequence lengths $\mathcal{N}$ \textbullet }
\label{sequence_length_ablation_section} In particular, we demonstrate  that the experiments in \ref{few_token_10_experiments}, which use sequence length $\mathcal{N}=3$, do not imply any loss of generality. We demonstrate this by virtue of ablation studies with varying sequence lengths, shown in Figure \ref{ablation_graph_on_length_of_path}.  Specifically, we examine the effect of increasing the path length between the starting token and the cycle token from $\mathcal{N}=4$ to $\mathcal{N}=64$. When keeping the default embedding dimension of 36 (as in Figure \ref{few_token_10_experiments}), generalization slows as sequence length increases. However, increasing the embedding dimension to 256 enables generalization across all token lengths within significantly fewer optimization steps. These findings align with prior work on the boundary between grokking and memorization in language models \cite{liu2022towards}, suggesting that varying sequence lengths would only introduce an added step of hyperparameter tuning.

In summary, the deterministic few-token experiments of Section \ref{fewtoken_section} provide initial evidence that self-referencing causal cycles, created by cycle tokens, enable models to effectively recall left-hand sequences and mitigate the reversal curse, as was outlined theoretically in Section \ref{formalizing_cycle_tokens}.

\subsection{Stochastic Few-Token \textsc{ReCall}}
\label{stochastic_experiments_section}

The experiments in Section \ref{fewtoken_section} assumed that each cycle token-sequence was followed by a unique token-sequence. For instance, in the single-token \texttt{Hyperlink Composability} experiment (Table \ref{table_for_few_token}), each $e1$ was succeeded by a unique $e4$ token.
\begin{table}[H]
\centering
\resizebox{\linewidth}{!}{%
    \begin{tabular}{c c c c} 
    \toprule 
    Experiment & Memorized Sequence & Reversal Path & Viable \\
    \midrule 
    \multirow{2}{*}{\makecell{Direct \\ Stochasticity}} & (e1, \{e2$_i$
    \}$_{i=1}^n$, e3, e1) & e3 $\rightarrow$ e1 $\rightarrow$ e2$_{i}$ & \includegraphics[height=1em]{images/bluecheckmark_thick.png} \\
    \cmidrule(lr){2-4}
    & (e1, \{E2$_i$\}$_{i=1}^n$, E3, e1) & E3 $\rightarrow$ e1 $\rightarrow$ E2$_i$ & \includegraphics[height=1em]{images/bluecheckmark_thick.png} \\
    \midrule 
    \multirow{2}{*}{\makecell{Hyperlink \\ Stochasticity}}& \makecell{(\{e5$_i$\}$_{i=1}^n$, e3, e1, \{e4$_i$\}$_{i=1}^n$)  \\ and (e0, e1, e2, e3)} & e2 $\rightarrow$ e3 $\rightarrow$ e1 $\rightarrow$ e4$_i$ & \includegraphics[height=1em]{images/bluecheckmark_thick.png} \\
    \cmidrule(lr){2-4}
    & \makecell{(\{E5$_i$\}$_{i=1}^n$, E3, e1, \{E4$_i$\}$_{i=1}^n$)  \\ and (E0, e1, E2, E3)} & E2 $\rightarrow$ E3 $\rightarrow$ e1 $\rightarrow$ E4$_i$ & \includegraphics[height=1em]{images/bluecheckmark_thick.png} \\
    \bottomrule  
    \end{tabular}
}
\caption{Experiment capturing self-referencing causal cycle path under stochastic setting. The figure illustrates how stochastic conditions affect the model's recall of a target token from a candidate set. During training, a range of valid candidates is presented to the model. In the case of direct stochasticity, the candidate set is denoted as ${e2}_{i=1}^n$, where multiple valid values for $e2$ can be sampled. During testing, the objective is to determine whether the model can correctly retrieve a specific target token $e2_i$ from the indexed set $[1, n]$. This experiment highlights the model's handling of ambiguity introduced by stochastic sampling scenarios.}
\label{table_for_stochasticity}
\end{table} In practice, cycle token-sequences often appear multiple times with varying subsequent tokens. For instance, a poem's title may recur across a Wikipedia page in different contexts (Section \ref{paragraph_section}), introducing a \textit{candidate set} of possible left-hand completions, as described in Section \ref{mathematical_rigor_section}. To investigate how cycle tokens probabilistically select from a candidate set, we designed a series of experiments, building on the experimental settings of Section \ref{fewtoken_section}.

In Table \ref{table_for_stochasticity}, the \texttt{Direct Stochasticity} experiment expands upon the \texttt{Baseline} few-token experiment by introducing a candidate set $\{e2_i\}_{i=1}^n$. For each $(e1,e2_i,e3)$ combination, we generate $n$ possible $e2_i$ tokens. For example, if $n=3$, then a fixed $e1$ sampled from $[1,100]$ and $e3$ from $[401,500]$ would recur in  3 samples with 3 distinct $e2$ values from $[101,400]$. The same reasoning applies to sequences, where a single $E2$ maps to $n$ distinct sequences $\{E2_i\}_{i=1}^n$. Here, $n$ represents the number of candidates, distinct from the sequence length $\mathcal{N}$.

\begin{figure}[!t]
\centering
\includegraphics[width=1.0\linewidth]{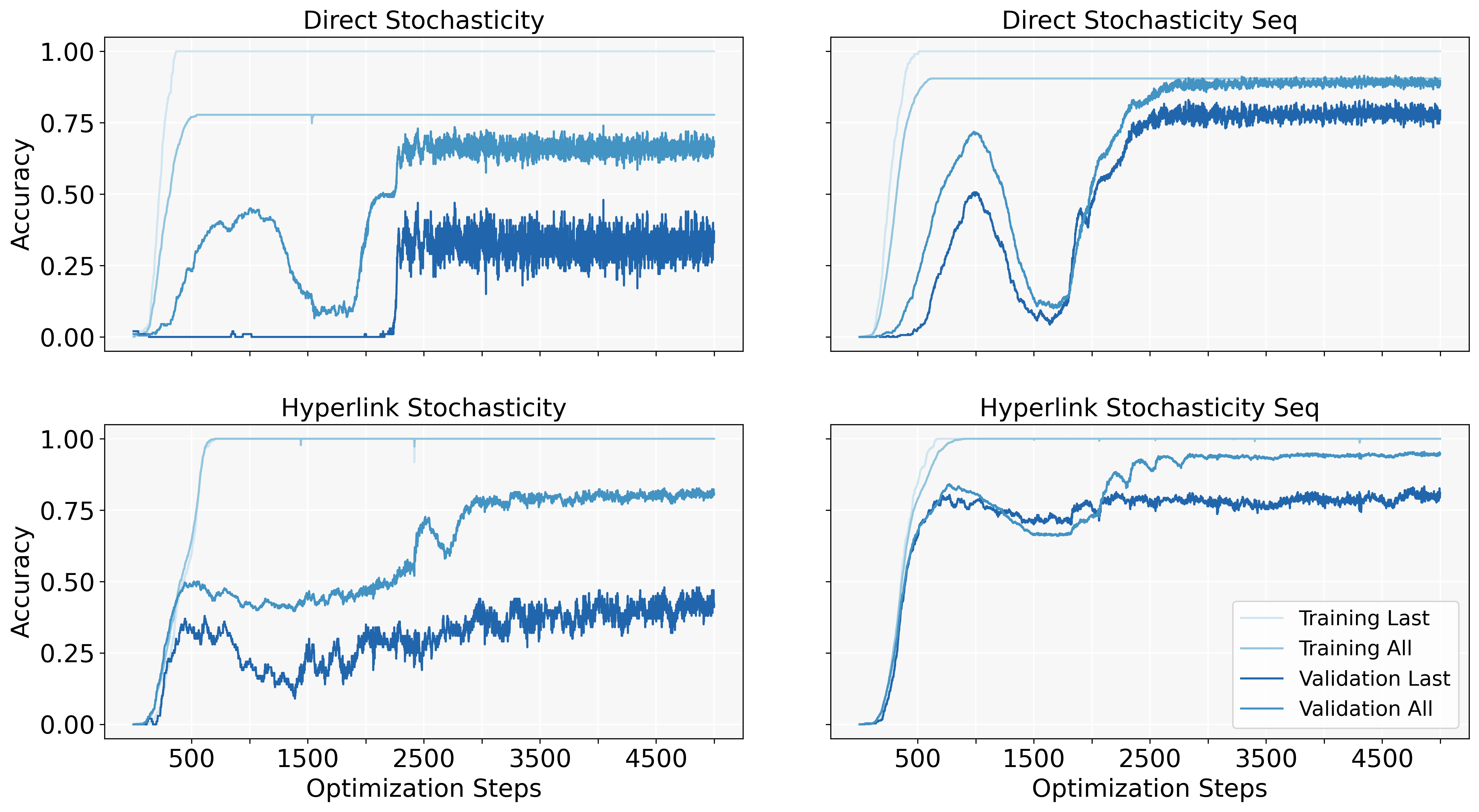}
\caption{Accuracy curves for four stochastic few-token tasks: The left column shows token-level tasks, and the right column shows sequence-level counterparts at candidates size $n=3$. The curves show accuracy on both the final token and all tokens, demonstrating the model’s ability to resolve specific target tokens despite stochastic variation during training.}
\label{stochastic_few_token_experiments}
\end{figure} \begin{figure}[!t] 
\centering
\includegraphics[width=1.0\linewidth]{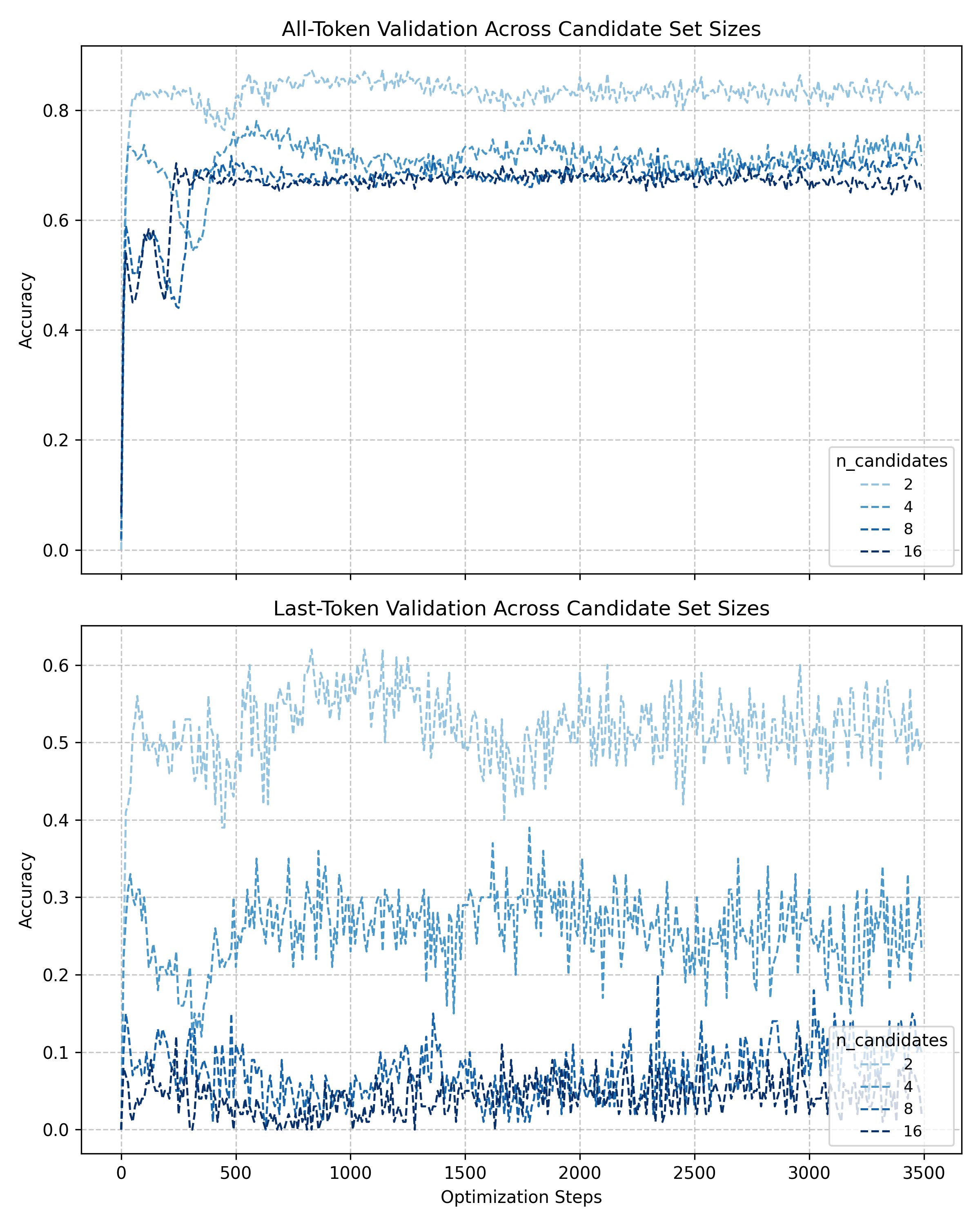}
\caption{As the candidate set size ($n$) increases, the accuracy of selecting the next token following the cycle token decreases proportionally, following a $\frac{1}{n}$ pattern. This progression is not only \textit{natural} but also \textit{preferable} to the alternative $\frac{1}{\mathcal{V}}$, where $\mathcal{V}$ is the vocabulary size and $\mathcal{V} \gg n$. Our \textsc{ReCall}-aware prompting strategy aligns with a language model's ability to \textit{enumerate} all $n$ options from the candidate set.}
\label{ablation_num_candidates_token_case}
\end{figure}
Similarly, the \texttt{Hyperlink Stochasticity} experiment expands on the \texttt{Hyperlink Composability} case by introducing $n$ candidates for both $e4$ and $E4$. The cycle token continues to facilitate a ``backward jump'', with additional candidates for $e5$ and $E5$ that do not disrupt the causal path.

\textbf{ Candidates can be targeted \textbullet} Figure \ref{stochastic_few_token_experiments} (left) provides evidence for this, confirming that the model can access any element within the candidate set.  For the single-token cases in both experiments, shown in the left-hand subplots, we set $n=3$ and observe the model retrieving one of three $e2_i$ tokens, with an expected accuracy of $\approx$ 33\% ($\frac{1}{3}$), consistent with uniform sampling. Sequence cases, shown in the right-hand subplots, follow a similar pattern. Notably, accuracy exceeds $\frac{1}{3}$ in the sequence case, since correctly predicting the first token ensures the correctness of subsequent tokens, increasing peak validation accuracy.

\textbf{Targeting is consistent across different candidate set sizes $n$ \textbullet} Specifically since $n$ was initially fixed in our experiments, we also explored the effect of varying $n$. Figure \ref{ablation_num_candidates_token_case} (left) illustrates a key observation, whereby we have an inverse relationship between the number of candidates and the accuracy of predicting the immediately succeeding token. For instance, under single-token \texttt{Direct Stochasticity}, 2 candidates yield $\frac{1}{2}$ accuracy, 4 candidates yield $\frac{1}{4}$, and so on.
 
\begin{tcolorbox}[colback=gray!10, colframe=black, sharp corners, boxrule=0.2mm, top=2mm, bottom=2mm, before skip=4mm, after skip=4mm, width=\linewidth]
\fontsize{9pt}{12pt}\selectfont
\noindent \textbf{The Reliability of The Candidate Set.} \textit{Even if the model does not select a target candidate from the candidate set, the target will still be one of the remaining $n-1$ candidates.} Consider the case of $n = 3$. The model always achieves 33\% accuracy in predicting the next token after the cycle token. That is, for each $e_1$, any $e_{2_i}$ is predicted with probability $\frac{1}{n}$. Since the candidate set contains only three possible $e_{2_i}$ values, the only other tokens in the vocabulary to be assigned non-zero probability are the other two $e_{2_i}$ from the candidate set. Since the model does not favor any particular $e_{2_i}$ due to the uniform distribution of predictions, the remaining $\frac{(n-1)}{n}$, two-thirds in this example, of the \textit{probability mass} must be split equally across the other two candidates. Otherwise, the model would retrieve target candidates with a probability of less than $\frac{1}{n}$. More generally, for any given cycle token, the model assigns non-zero probabilities to each candidate, and their probabilities are balanced. Thus, any of the $n$ candidates will always appear within the top-$n$ predictions, and by extension, must be included in the candidate set, even when they are not selected by the model.
\end{tcolorbox}

\begin{figure*}[!t]
\centering
\includegraphics[width=1.0\linewidth]{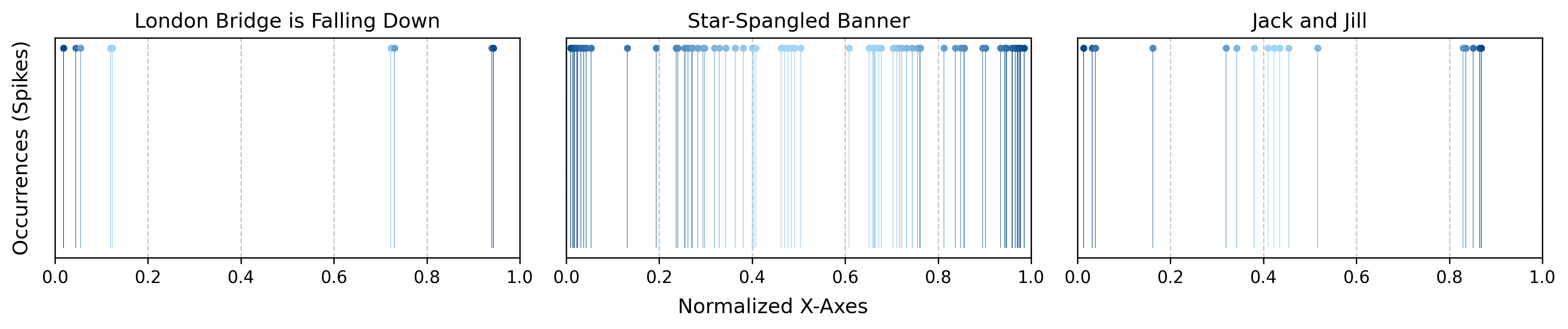}
\captionsetup{skip=0.2em}
\caption{Spatial distribution of cycle token-sequences across key-writing texts. The lengths of each key-writing, one per subplot, are normalized to a 0-1 scale, with each vertical line representing a recurrence of the cycle token-sequence.  For ease of observability, the colors at the beginning and end of the text are shaded darker blue. The visualization highlights how clusters of cycle token-sequences act as hyperlinks in pretraining data, enabling models to reference prior information effectively. Derived from the titles of key-writings, these cycle token-sequences facilitate the 'jump backward' mechanism of self-referencing causal cycles, which LLMs leverage as natural aids for robust memory retrieval.}
\label{pagaraph_historgram_2}
\end{figure*} 

Stochastic self-referencing causal cycles provide key insights into the reversal curse and tailored prompting strategies. By decoupling candidate set generation from selection, models better leverage the causal pathways created by cycle tokens for next-token prediction.

\subsection{\textsc{ReCall} in Pretraining Corpuses}
\label{paragraph_section}

\subsubsection{Observing Cycle Token-Sequences} Viewing LLMs as dynamic knowledge repositories, we observe self-referencing causal cycles in widely recognized texts, which we refer to as \textit{key writings}. These recurring phrases act as conceptual hyperlinks, guiding the retrieval of stored information. This hyperlinking behavior, while often overlooked, represents a core mechanism by which LLMs bridge long-range dependencies in text. Our key writings include timeless poems, iconic speeches, and universally familiar nursery rhymes, forming the backbone of cultural memory. 

To curate our dataset, we queried ChatGPT for examples of popular texts and refined the selection to 50 key writings easily recognizable to those familiar with English literature. A detailed list, along with relevant weblinks, is provided in Appendix \ref{sample_popular_texts}. We cross-referenced this list with a Llama 3 405B model to verify the texts were part of its pretraining data. This verification was conducted offline to ensure no reliance on live queries. Additionally, we analyzed associated webpages to understand how key writings are embedded in real-world corpora \cite{brown2020language,touvron2023llama,wang2021gptj,gao2021pile}. Examples are shown in Figures \ref{paragraph_texts_1} and \ref{paragraph_texts_2} (in the Appendix).

\begin{figure}[H]
\centering
\includegraphics[width=1.0\linewidth]{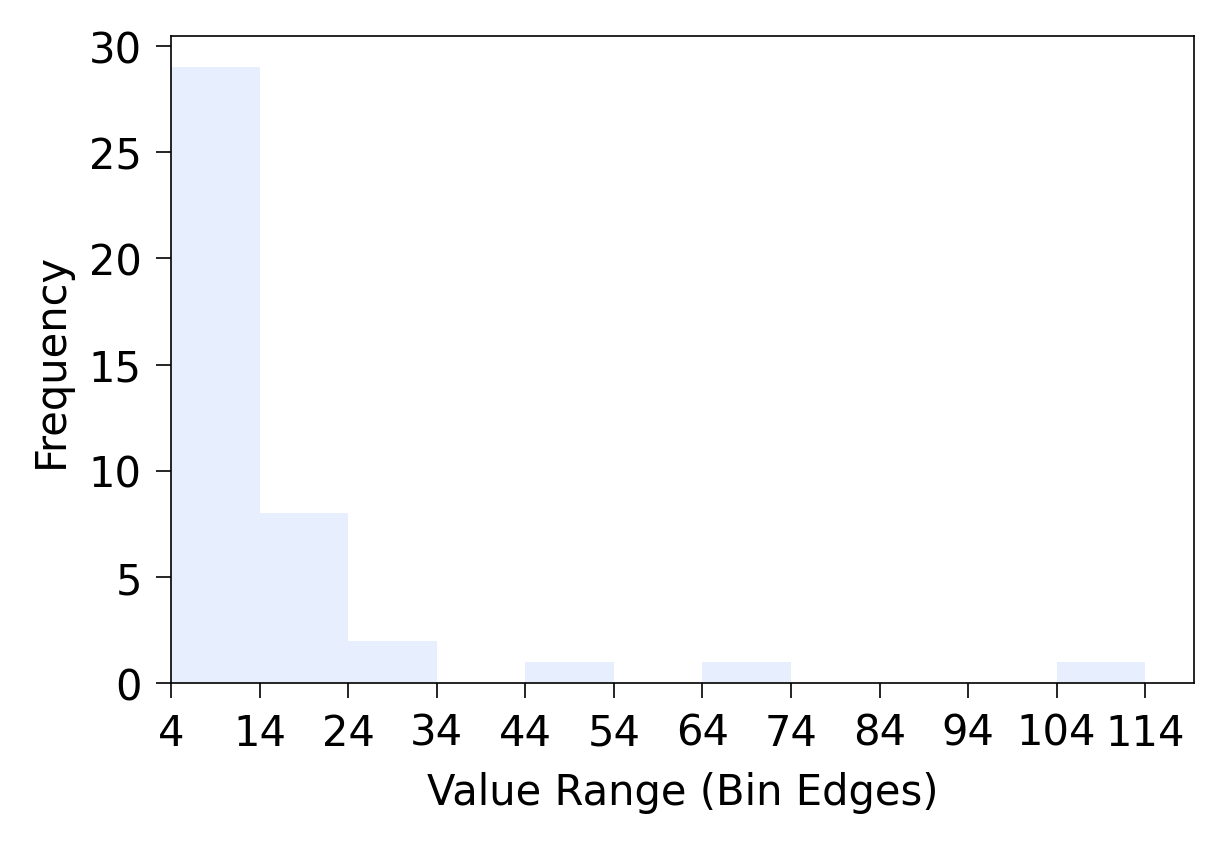} 
\caption{Frequency of cycle token-sequences across pretraining corpora for selected key writings. The bar chart illustrates the number of times cycle token-sequences, such as titles, appear as recurring motifs within corresponding webpages. Even if some occur only a few times per webpage, this suffices to create natural self-referencing causal cycles, enabling LLMs to bridge distant parts of text and achieve robust memory recall during next-token prediction.}
\label{pagaraph_historgram_1}
\end{figure}

We analyzed the frequency and distribution of cycle token sequences across relevant webpages, treating each key writing's title—or a subsequence of it—as a cycle token. These sequences act like cross-references in a library catalog, linking disparate sections of text and enabling efficient retrieval of distant information. Figure \ref{pagaraph_historgram_1} illustrates how often these tokens recur, with phrases like ``Star-Spangled Banner'' appearing 73 times in its corresponding Wikipedia article for the U.S. anthem. Figure \ref{pagaraph_historgram_2} highlights their distribution, showcasing the extensive causal pathways that cycle tokens create, allowing backward retrieval without breaking left-to-right prediction flow. 

\subsubsection{Utilizing \textsc{ReCall}} 
In a library, readers navigate between sections by following references or repeated titles. Similarly, cycle tokens enable language models to ``jump'' across text sections, retrieving relevant information even when it is stored far from the original query. Without these natural hyperlinks, models often struggle to recall preceding information accurately. One example is the ``preceding line problem,'' adapted from \cite{golovneva2024reverse}. For instance, GPT-4 correctly identifies the line following ``Gave proof through the night that our flag was still there'' in the U.S. national anthem as ``O say does that star-spangled banner yet wave.'' However, when asked for the preceding line, it consistently returns incorrect responses, such as ``O'er the ramparts we watch'd, were so gallantly streaming?'' This issue persists across models, including Llama-3-405B \cite{dubey2024llama}, Google Gemini \cite{team2023gemini}, and Claude 3.5 Sonnet \cite{anthropic2024claude35}, despite various prompting strategies like step-by-step reasoning, process of elimination, and structured planning.

\begin{figure}[H]
\centering
\includegraphics[width=1.0\linewidth]{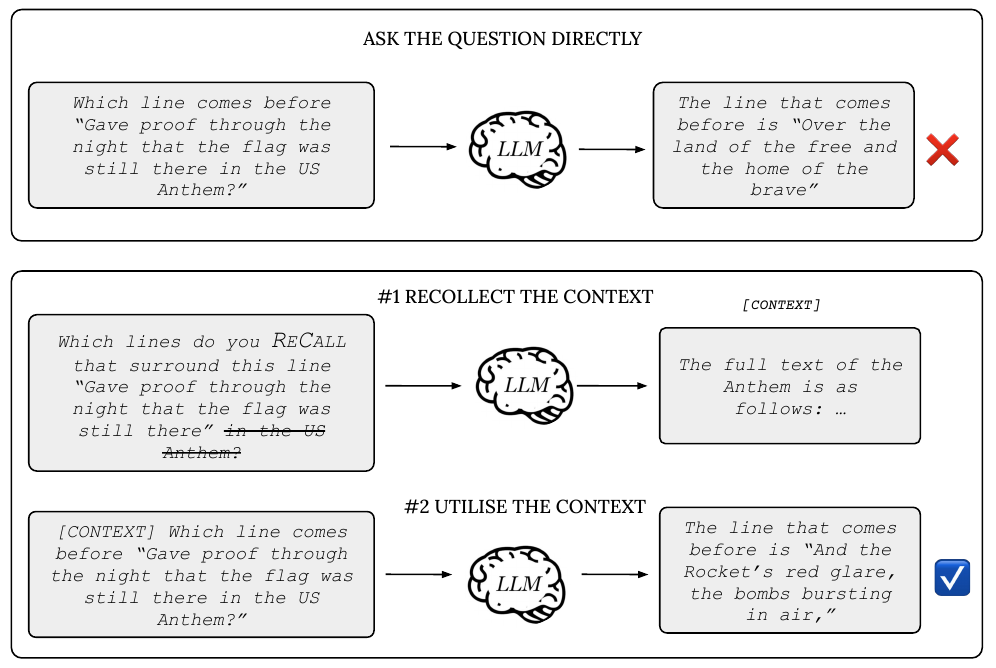}
\caption{Two-step \textsc{ReCall}-aware prompting to resolve the U.S. National Anthem's preceding line problem. The first step involves recollecting all relevant context surrounding the given line using a broad query. The second step utilizes the retrieved context to correctly identify the preceding line. This approach leverages self-referencing causal cycles to ensure the model explores all causal pathways, overcoming the reversal curse. It is effective in 100\% of the key-writings we test.}
\label{two_step_process}
\end{figure}

To address this limitation, we propose a two-step \textsc{ReCall}-aware prompting strategy. Unlike conventional prompts, which often fail due to the next-token prediction bias, \textsc{ReCall}-aware prompting explicitly retrieves context before answering. This strategy mandates exploring all causal pathways, including self-referencing causal cycles, which are, strictly speaking, causal pathways, and serve as natural hyperlinks within the training data. The first step involves asking the model to recall everything it knows about the token-sequence of interest. For instance, when querying the line ``X'' in a key-writing, we ask, \texttt{``Tell me the lines surrounding this line `X'.''} For verbose models, a broader query such as \texttt{``Tell me everything you know about this line `X'''} often suffices to retrieve the necessary context. Once the model outputs a candidate set of surrounding lines, the second step extracts the correct answer through in-context learning. This process circumvents the inherent expediency bias of next-token prediction. Figure \ref{two_step_process} demonstrates how this approach resolves the preceding line problem in the U.S. national anthem. Vitally, this approach is effective in 100\% of our key-writings for GPT-4o (2024-12-23) and LlaMA-3.3-70B. For any given complete and intelligible sentence in the key-writing, we are able to retrieve the preceding one.
 
In summary, the \textsc{ReCall}-aware prompting process involves two steps:
\begin{enumerate}
 \setlength\itemsep{-.1 em}
    \item Recollect the context: Prompt the model to provide a candidate set of answers by leveraging self-referencing causal cycles.
    \item Utilize the context: Instruct the model to analyze its own outputs and extract the correct answer through in-context reasoning.
\end{enumerate}

\section{Conclusions} 

The reversal curse is a well-documented challenge in generative language models, where the model struggles to predict preceding tokens based on succeeding tokens. While prior work has primarily focused on data augmentation or architectural changes to address this limitation, we propose an alternative perspective: language models, much like libraries, may already possess latent mechanisms to cross-reference token sequences within their pretraining data.

In this work, we introduce the concept of self-referencing causal cycles (abbreviated \textsc{ReCall}), which act as natural hyperlinks in a model's memory, allowing it to retrieve left-hand tokens from right-hand tokens. We demonstrate this concept at an axiomatic level through controlled few-token experiments on a small transformer model. The flexibility of these cycles under stochastic conditions suggests that \textsc{ReCall} mechanisms can scale to large language models. Notably, we find experimentally that self-referencing cycles emerge naturally from repeated token patterns in pretraining corpora and can enable models to overcome the reversal curse without requiring additional modifications. By leveraging these cycles, language models are able to overcome the limitations of autoregressive text generation and produce more reliable responses to prompts affected by the reversal curse.

\section*{Limitations}

While this study demonstrates the potential of self-referencing causal cycles to mitigate the reversal curse, there are several limitations to consider. Our experiments are conducted in controlled settings with simplified token sequences. However, the performance of autoregressive models deployed in real-world applications may be influenced by additional factors, such as retrieval-augmented generation or web search. Further interpretability techniques may be required to precisely attribute parametric information retrieval to specific cycle tokens in the pretraining data. This poses a non-trivial challenge, as larger models often utilize fully or partially closed-source training data, and extracting pretraining data from the models themselves is intentionally designed to be difficult to maintain privacy and security.

\section*{Acknowledgments}

\bibliography{acl_latex}

\appendix
\section*{Appendix}

\section{Sample Popular Texts}
\label{sample_popular_texts}

\begin{enumerate}
\setlength{\itemsep}{-0.3em}

    \item \href{https://en.wikipedia.org/wiki/The_Star-Spangled_Banner#Lyrics}{Star-Spangled Banner (U.S. Anthem)} 
    \item \href{https://en.wikipedia.org/wiki/London_Bridge_Is_Falling_Down#Lyrics}{London Bridge is falling down}
    \item \href{https://en.wikipedia.org/wiki/Jack_and_Jill#Text}{Jack and Jill}
    \item \href{https://en.wikipedia.org/wiki/Baa,_Baa,_Black_Sheep#Modern_version}{Baa Baa Black Sheep}
    \item \href{https://en.wikipedia.org/wiki/Twinkle,_Twinkle,_Little_Star#Lyrics}{Twinkle Twinkle Little Star}
    \item \href{https://www.npr.org/2010/01/18/122701268/i-have-a-dream-speech-in-its-entirety}{"I have a dream" by Martin Luther King Jr.}
    \item \href{https://en.wikipedia.org/wiki/Humpty_Dumpty#Lyrics_and_melody}{Humpty Dumpty}
    \item \href{https://en.wikipedia.org/wiki/Jingle_Bells}{Jingle Bells}
    \item \href{https://en.wikipedia.org/wiki/Three_Blind_Mice#Lyrics}{Three Blind Mice}
    \item \href{https://en.wikipedia.org/wiki/The_Twelve_Days_of_Christmas_(song)#Lyrics}{The Twelve Days of Christmas}
    \item \href{https://en.wikipedia.org/wiki/Fee-fi-fo-fum}{Fee-fi-fo-fum}
    \item \href{https://en.wikipedia.org/wiki/Three_Little_Kittens#Text}{Three Little Kittens}
    \item \href{https://genius.com/Christmas-songs-frosty-the-snowman-lyrics}{Frosty The Snowman}
    \item \href{https://en.wikipedia.org/wiki/Old_soldiers_never_die}{Old Soldiers Never Die}
    \item \href{https://en.wikipedia.org/wiki/Do_not_go_gentle_into_that_good_night}{Do not go gentle into that good night}
    \item \href{https://www.nps.gov/articles/sojourner-truth.htm}{Ain’t I A Woman by Sojourner Truth}  
    \item \href{https://en.wikipedia.org/wiki/It%27s_Raining,_It%27s_Pouring}{It's Raining, It's Pouring}
    \item \href{https://www.poetryfoundation.org/poems/52829/a-dream-within-a-dream}{A Dream Within A Dream}
    \item \href{https://en.wikipedia.org/wiki/This_Old_Man#Variations}{This Old Man}
    \item \href{https://en.wikipedia.org/wiki/Peter_Piper#Lyrics}{Peter Piper}
    \item \href{https://en.wikipedia.org/wiki/Old_Mother_Hubbard#Words}{Old Mother Hubbard}
    \item \href{https://en.wikipedia.org/wiki/Matthew,_Mark,_Luke_and_John#Lyrics}{Matthew, Mark, Luke and John}
    \item \href{https://en.wikipedia.org/wiki/Little_Bo-Peep#Lyrics_and_melody}{Little Bo-Peep}
    \item \href{https://en.wikipedia.org/wiki/There_Was_an_Old_Woman_Who_Lived_in_a_Shoe#Lyrics}{There was an Old Woman who lived in a shoe}
    \item \href{https://en.wikipedia.org/wiki/The_Grand_Old_Duke_of_York#Words}{The Grand Old Duke of York}
    \item \href{https://en.wikipedia.org/wiki/Solomon_Grundy_(nursery_rhyme)#Lyrics}{Solomon Grundy (nursery rhyme)}
    \item \href{https://en.wikipedia.org/wiki/Star_Light,_Star_Bright#Lyrics}{Star Light, Star Bright}
    \item \href{https://en.wikipedia.org/wiki/John_Jacob_Jingleheimer_Schmidt#Lyrics_and_melody}{John Jacob Jingleheimer Schmidt}
    \item \href{https://en.wikipedia.org/wiki/Rock-a-bye_Baby#Words}{Rock-a-bye Baby}
    \item \href{https://en.wikipedia.org/wiki/Pussy_Cat_Pussy_Cat#Lyrics_and_melody}{Pussy Cat Pussy Cat}
    \item \href{https://en.wikipedia.org/wiki/The_Queen_of_Hearts_(poem)#Synopsis_and_structure}{The Queen of Hearts}
    \item \href{https://www.npr.org/2008/11/05/96624326/transcript-of-barack-obamas-victory-speech}{Barack Obama's Victory Speech 2008 (\textsc{ReCall} `Yes, we can')} 
    \item \href{https://en.wikipedia.org/wiki/Little_Miss_Muffet}{Little Miss Muffet}
    \item \href{https://en.wikipedia.org/wiki/Two_Little_Dickie_Birds#Lyrics}{Two Little Dickie Birds}
    \item \href{https://en.wikipedia.org/wiki/Little_Arabella_Miller#Lyrics}{Little Arabella Miller}
    \item \href{https://en.wikipedia.org/wiki/Little_Robin_Redbreast#Lyrics}{Little Robin Redbreast}
    \item \href{https://en.wikipedia.org/wiki/Doctor_Foster_(nursery_rhyme)#The_rhyme}{Doctor Foster} 
    \item \href{https://en.wikipedia.org/wiki/Pat-a-cake,_pat-a-cake,_baker%27s_man}{Pat-a-cake, pat-a-cake, baker's man }(\textsc{ReCall} Pat-a-cake)
    \item \href{https://en.wikipedia.org/wiki/Tom,_Tom,_the_Piper%27s_Son}{Tom, Tom, the Piper's Son}
    \item \href{https://en.wikipedia.org/wiki/Little_Jack_Horner}{Little Jack Horner}
    \item \href{https://en.wikipedia.org/wiki/99_Bottles_of_Beer}{"99 Bottles of Beer" by John Donne} 
    \item \href{https://en.wikipedia.org/wiki/Hickory_Dickory_Dock}{Hickory Dickory Dock} 
\end{enumerate}

\begin{figure*}[!t]
\centering
\includegraphics[width=0.8\linewidth]{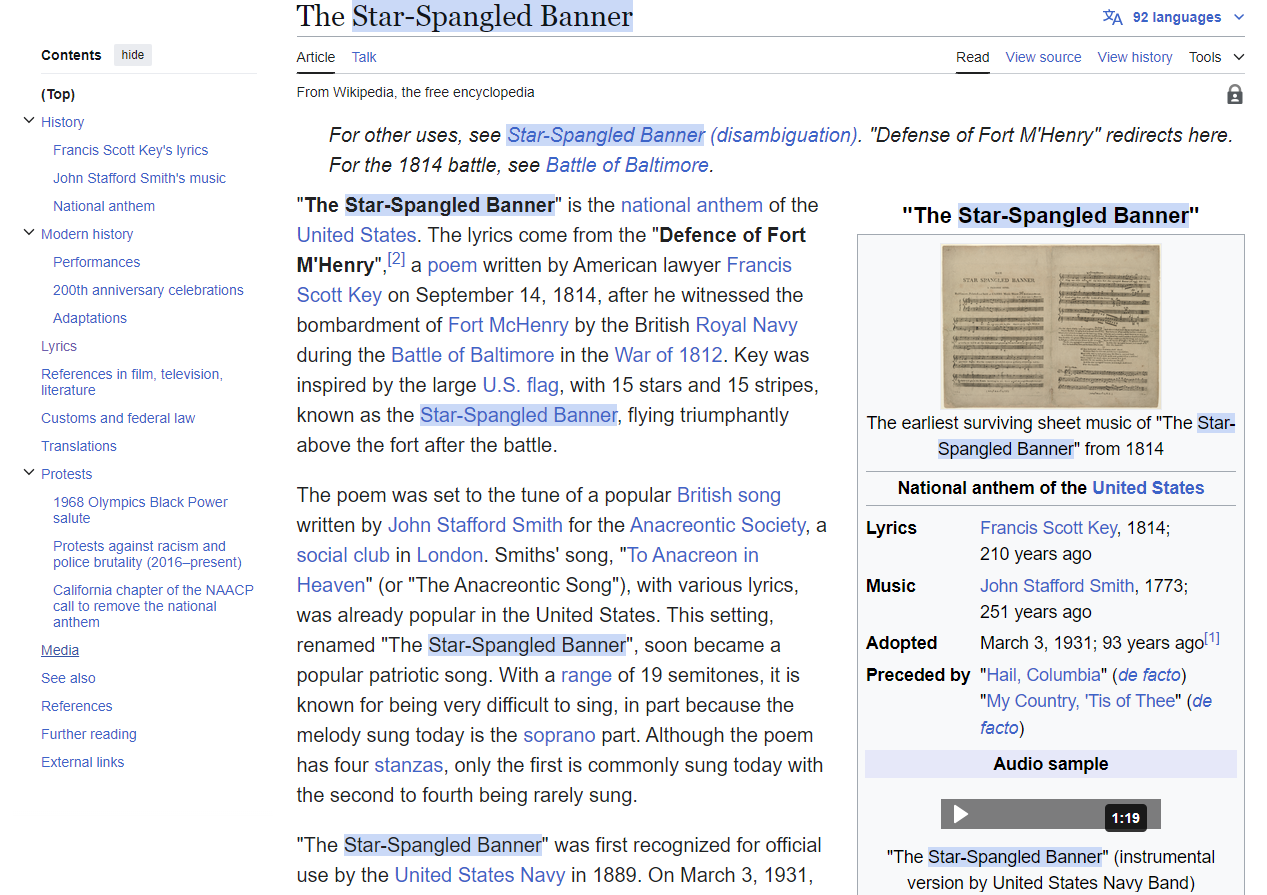}
\caption{An example of a webpage containing a key-writing about the U.S. National anthem's title. The text includes self-referencing causal cycles induced by the Anthem title. This figure demonstrates how the repeated phrases within the text create natural hyperlinks, enabling the model to retrieve contextual information in the case of the ``preceding line problem.''}
\label{paragraph_texts_1}
\end{figure*}  

\begin{figure*}[!t]
\centering
\includegraphics[width=0.8\linewidth]{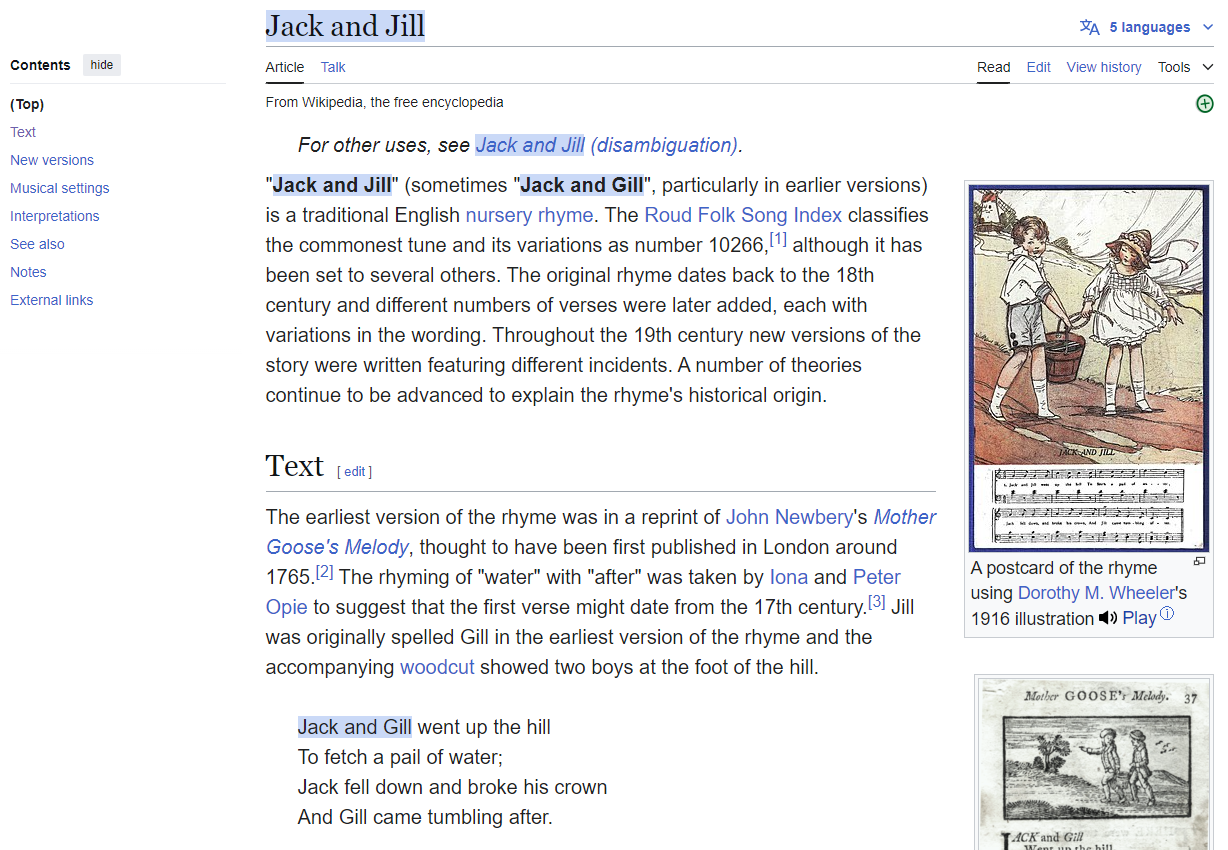}
\caption{An example of webpage containing a key-writing about the well-known poem ``Jack and Jill''. LLMs are able to bridge semantic gaps caused by slight variations in spelling, such as ``Jack and Jill'' vs. ``Jack and Gill.'' However, this invariance enhances the robustness of self-referencing causal cycles in real-world datasets.}
\label{paragraph_texts_2}
\end{figure*} 

\section{Computing a Causally Preceeding Sequence with an Autoregressive Model}
\subsection{The Argmax Over $\mathcal{S}$}
\label{background_probability} 

We shall illustrate how we arrive at Equation \ref{explained_in_appendix} from Equation \ref{original_equation} in Section \ref{mathematical_rigor_section}.

Firstly, as per Equation \ref{original_equation}, we have:
\[
S_r = \argmax_{s \in \mathcal{S}} P_{\mathcal{M}}(s | S_l).
\]

The true probability distribution of a left-hand sequence seen in the training data, given a right-hand sequence, is given by:
\begin{equation}
P(s | S_r) =\displaystyle \frac{ P(S_r | s) P(s) }{P(S_r)}
\end{equation}
as per Bayes rule. Equivalently, we have that: 
\begin{align}
\argmax_{s \in \mathcal{S}} P(s | S_r) &= \argmax_{s \in \mathcal{S}} \frac{ P(S_r | s) P(s) }{P(S_r)} \\
&= \argmax_{s\in \mathcal{S}}\displaystyle \frac{P(S_r|s)P(s)}{C} \\
&= \argmax_{s \in \mathcal{S}}\displaystyle P(S_r|s)P(s),
\end{align} 
where $C$ is a constant, indicating that ${P(S_r)}$ does not affect the final $\argmax$.  

If the LLM is a perfect model of the true data distribution (i.e., a perfect library), 
\begin{equation}
\forall s\in \mathcal{S}: P(s) = P_{\mathcal{M}}(s).
\end{equation}

Thus we have:
\begin{equation}
\argmax_{s \in \mathcal{S}} P(s | S_r) = \argmax_{s \in \mathcal{S}} P_{\mathcal{M}}(S_r|s)P_{\mathcal{M}}(s),
\end{equation}
and finally, if $S_l$ is indeed the most natural preceding sequence to $S_r$ among all possible sequences, then: 
\[
S_l = \argmax_{s \in \mathcal{S}} P_{\mathcal{M}}(S_r|s)P_{\mathcal{M}}(s).
\]
Which is Equation \ref{explained_in_appendix}. Here $P_{\mathcal{M}}(s)$ is given by the model as the frequency of the token-sequence $s$ in the pre-training corpus. For rare sequences, $P_{\mathcal{M}}(s)$ should be roughly small, and uniform, while for sequences that are popular within the dataset (see curse of popularity \cite{takahashi2024curse}),  $P_{\mathcal{M}}(s)$ will have a large value.

In either case, we can use the LLM to easily compute $P_{\mathcal{M}}(s)$ as a correction factor for  $P_{\mathcal{M}}(S_r|s)$, indicating an autoregressive LLM should be able to compute the most likely left-hand sequence given a right-hand sequence, even though it was trained to give the right-hand part after the left-hand part.

\subsection{The Candidate Set $S_{l_c} \subset \mathcal{S}$}
\label{candidate_set}
One significant challenge in utilizing self-referencing causal cycles lies in efficiently computing the $\argmax$ over possible left-hand sequences $s \in \mathcal{S}$. A brute-force approach, iterating through all candidates $s \in \mathcal{S}$, is computationally infeasible. For a fixed vocabulary size $v$, the total number of $k$-length sequences is $v^k$, which grows exponentially as $k$ increases.

To address this, we aim to construct a smaller candidate set $S_{l_c} \subset \mathcal{S}$, with $|S_{l_c}| \ll |\mathcal{S}|$, that can serve as a proxy for the full set $\mathcal{S}$. By narrowing down the candidate set intelligently, we can perform the necessary $\argmax$ computation efficiently. Specifically, at each iteration, we pair a candidate $s \in S_{l_c}$ with $S_r$ and evaluate the likelihood of observing $\left[s, S_r\right]$ in context. Formally, this is expressed as:

$$S_l= \argmax_{s \in S_{l_c}}P_{\mathcal{M}}(S_r|s)P_{\mathcal{M}}(s)$$

If $S_{l_c}$ is constructed effectively, the $\argmax$ computed over $S_{l_c}$ will match the $\argmax$ over the entire set $S$.

\section{Examples are Insufficient for Reversal}
\label{generalization_candidate_set} 

We conducted a token experiment using an 8-layer transformer designed to memorize simple token-sequences of the form $(e,r,f)$. Each sample was constructed by randomly selecting $e$ from $[1,10000]$ and $f$ from $[10001,20000]$, pairing with one of two relation tokens: $r = 20001$ or its inverse $r' = 20002$. The notation $(e,r,f)$ stands for entity, relation, feature, and is analogous to the $(s,r,o)$ structure commonly used in works such as \cite{takahashi2024curse}, where $e$ and $f$ correspond to subject and object entities, respectively. Importantly, each $e$ is paired uniquely with an $f$, resulting in $20000$ unique entity-feature pairings, appearing in one of four possible configurations: \\

\vspace{-.5cm} 
\begin{description}[labelindent=2cm]
\setlength\itemsep{-.1 em}
    \item $(F,e,r,f)$ — (1)  
    \item $(F,f,r',e)$ — (2)  
    \item $(R,f,r,e)$ — (3)  
    \item $(R,e,r',f)$ — (4)  
\end{description}

\begin{table}[!h]
\small
\centering
\resizebox{1.0\linewidth}{!}{
\setlength{\tabcolsep}{0.5em}
\renewcommand{\arraystretch}{1.8}
\begin{tabular}{|c|c|c|c|} 
\hline \multirow{1}{*}{Train} & \makecell{(1) + \\ 1/2 of (2)} &  \makecell{ (1) + 1/2 of (2) \\  + (3) + 1/2 of (4)}  &   \makecell{ (1) + 1/2 of (2) \\ + 1/2 of (3) } \\
\cline{1-4} 
\multirow{1}{*}{Test} & 1/2' of (2)&  1/2' of (2) & 1/2' of (2)  \\ \cline{1-4}
\multirow{1}{*}{Score} & 0 & 100 & 0 \\ \cline{1-4}
\multirow{1}{*}{Type} & Standard  & Reverse Training & Generalization \\ 
 \hline  
\end{tabular}}
\caption{Comparison of training strategies to achieve reversal via examples.}
\label{generalisation_table}
\end{table}

For instance, $e$ could represent names of people, $f$ could represent names of objects, and $r$ and $r'$ could represent actions such as `kicks' and `is kicked by,' respectively. We also introduce special forward ($F$) and backward ($R$) directional tokens to distinguish between left-to-right and reverse-direction data. Our goal was to evaluate the notion of generalization proposed in \cite{golovneva2024reverse} using a variant of this experiment, focusing on whether the model can generalize right-to-left relations from left-to-right ones.

Consider the three training configurations summarized in Table \ref{generalisation_table}. In standard training, the model is shown all $(e,r,f)$ relations and the first half of the $(f,r',e)$ relations. It is then tested on the second half (denoted 1/2', pronounced `half-prime') of the $(f,r',e)$ relations. Here testing entails inputting the first two tokens and trying to predict the last correctly. Although the $e$-and-$f$ token pairs in the second half of the $(f,r',e)$ samples directly correspond to those in the second half of the $(e,r,f)$ samples from training, the model struggles to infer the reverse relations, achieving a score of 0 in the second column. For instance, given `John kicks a ball,' it fails to deduce `a ball is kicked by John.' This is to be expected, due to the reversal curse.

In reverse training, as described in \cite{golovneva2024reverse}, the model is trained on all $(e,r,f)$ relations and all $(f,r,e)$ relations, where the latter are simply the result of applying token permutation (all-token reversal) to the former. The model is then tested on the second half of $(f,r',e)$ relations, and achieves a perfect score of 100. This occurs because training on both $(e,r,f)$ and $(f,r,e)$ creates a direct causal link between $f$ and $e$, making it straightforward for the model to predict $e$ given $f$ (even when $r$ is substituted with $r'$). For example, if the model sees both `Mary kicks a bottle' and `a bottle kicks Mary' during training, it can easily complete `a bottle \textit{is kicked by}' with `Mary.'

However, reverse training does not constitute true generalization. For genuine generalization, the model must infer the second half of $(f,r',e)$ relations without having been trained on their corresponding $(f,r,e)$ relations. For instance, if the model is shown `John kicks a ball' and `a ball is kicked by John,' as well as `Mary kicks a bottle,' it will be able to complete `a ball \textit{is kicked by}' with `John.' However, will it be able to complete `a bottle \textit{is kicked by}' with `Mary'? This scenario is tested in the last column of Table \ref{generalisation_table}, where the model fails to achieve such generalization, scoring 0. This demonstrates that under next-token prediction loss constraints, generalization of reversal is unachievable, even with examples. This is consistent with findings in prior works in the literature.

\section{Experimental Settings}
\label{appendix_exp_settings}

For the deterministic few-token experiments of Section \ref{fewtoken_section}, we use a small ($\sim$90,000 parameters) decoder-based transformer model with 2 layers and 8 attention heads. The default model embedding dimension is 36, with the exception of the \texttt{Length-of-Path} and candidate set size experiments, for which it is increased to 256. Training is conducted using cross-entropy loss, which is well-suited for the next-token prediction problem in transformer models. We run multiple random seeds but observe no change in results due to the well-defined nature of the problem. The experiments, implemented in PyTorch and NumPy, were performed on an NVIDIA A100 GPU and trained with the Adam optimizer (learning rate: 0.001, batch size: 1024). Reproduction requires a compute budget of slightly over 1 hour (1 hour and 9 minutes in our rerun). For the stochastic case experiments in Section \ref{stochastic_experiments_section}, we maintain the same experimental settings, but with time required for reproduction increasing slightly to over 1.5 hours.

\end{document}